\documentclass{egpubl}
\usepackage{pg2024}

\SpecialIssuePaper         

\CGFStandardLicense

\usepackage[T1]{fontenc}
\usepackage{dfadobe}  

\usepackage{cite}  
\BibtexOrBiblatex
\electronicVersion
\PrintedOrElectronic
\ifpdf \usepackage[pdftex]{graphicx} \pdfcompresslevel=9
\else \usepackage[dvips]{graphicx} \fi

\usepackage{egweblnk} 
\usepackage{eucal}
\usepackage{amsmath}
\usepackage{amssymb}
\usepackage{soul}
\usepackage{microtype}
\usepackage{multirow} 
\usepackage{booktabs}
\usepackage{collect}
\usepackage{xspace}
\usepackage{dsfont}
\usepackage[colorlinks,linkcolor=blue]{hyperref}
\usepackage{xcolor}
\usepackage{hyperref}

\usepackage{xcolor}

\newcommand{\ie}{i.e.,\ }

\newcommand{\supp}[1]{\textcolor{red}{\hl{#1}}}

\renewcommand{\supp}[1]{#1}

\definecollection{mymaths}

\newcommand{\mymath}[2]{
    \newcommand{#1}{\TextOrMath{$#2$\xspace}{#2}}
    \begin{collect}{mymaths}{}{}{}{}
    #1
    \end{collect}
}

\mymath{\inputimage}{I}
\mymath{\numinputimages}{N}
\mymath{\mean}{\boldsymbol{\mu}}
\mymath{\covariance}{\Sigma}
\mymath{\opacity}{o}
\mymath{\gaussiancolor}{\mathbf{c}}
\mymath{\gaussianalpha}{\alpha}
\mymath{\pixelcolor}{C}
\mymath{\pixelposition}{\mathbf{x}}
\mymath{\projectedmean}{\hat{\mean}}
\mymath{\projectedcovariance}{\hat{\covariance}}
\mymath{\gaussiancolorhdr}{\mathbf{c}^{\text{HDR}}}
\mymath{\coc}{r}
\mymath{\focallength}{f}
\mymath{\aperture}{a}
\mymath{\focaldistance}{{d_f}}
\mymath{\depth}{d}
\mymath{\aperturegaussian}{\projectedcovariance^\aperture}
\mymath{\covariancedefocus}{{\projectedcovariance^{\text{DoF}}}}
\mymath{\gaussianalphadefocus}{\alpha^\text{DoF}}
\mymath{\normalizationfactor}{\beta}
\mymath{\pixelcolordefocushdr}{\pixelcolor_\text{HDR+DoF}}
\mymath{\pixelcolordefocusldr}{\bar{\pixelcolor}}
\mymath{\gammacorrection}{\gamma}
\mymath{\exposuretime}{t}
\mymath{\discrepancy}{\Delta}
\mymath{\blurdetector}{g}

\title{Cinematic Gaussians:\\\,Real-Time HDR Radiance Fields with Depth of Field}

\author[Wang et al.]
{\parbox{\textwidth}{\centering Chao Wang$^{1}$, Krzysztof Wolski$^{1}$, Bernhard Kerbl$^{2,3}$, Ana Serrano$^{4}$, Mojtaba Bemana$^{1}$,\\
Hans-Peter Seidel$^{1}$, Karol Myszkowski$^{1}$, and Thomas Leimk\"{u}hler$^{1}$
	}
	\\
	{\parbox{\textwidth}{\centering $^1$Max-Planck-Institut f\" ur Informatik, Germany \\
                $^2$Technische Universit\"at Wien, Austria\\
                $^3$Carnegie Mellon University, USA \\
                $^4$Universidad de Zaragoza, I3A, Spain
		}
	}
}

\begin{document}
 \teaser{
  \includegraphics[width=1\linewidth]{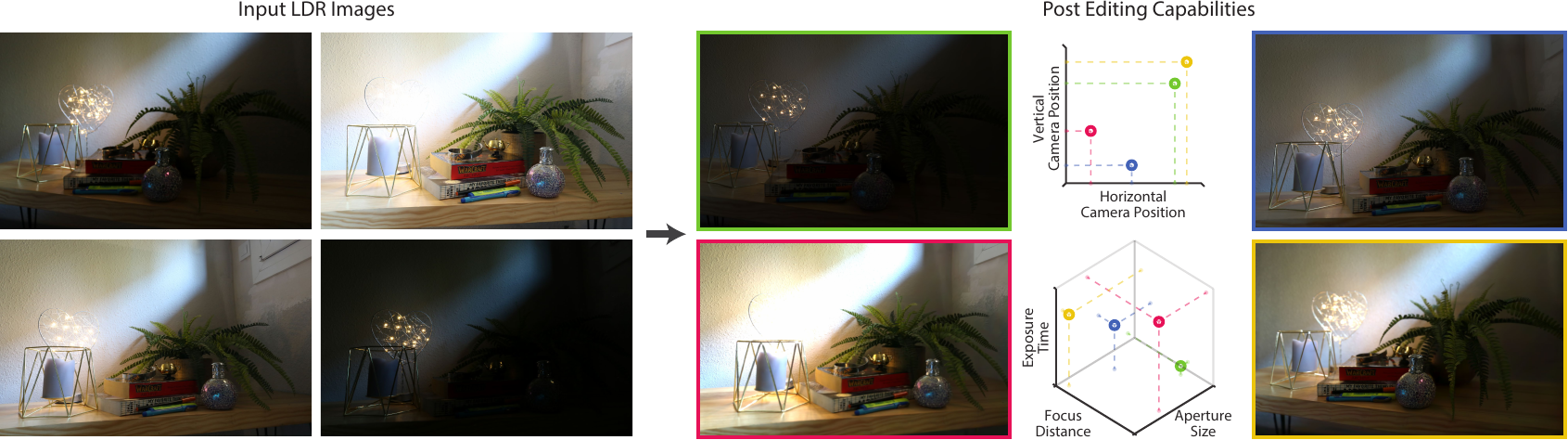}
  \centering
   \caption{Based on sparse images captured from various viewpoints and with different focus distances, apertures, and exposure times (\emph{left}), our method generates an HDR 3D Gaussian representation that can be rendered with arbitrary depth of field. Our representation supports rendering images in a six-dimensional space of camera and lens parameters (\textit{right, five dimensions are shown}).}
 \label{fig:teaser}
}


\maketitle


\begin{abstract} 

Radiance field methods represent the state of the art in reconstructing complex scenes from multi-view photos.
However, these reconstructions often suffer from one or both of the following limitations:
First, they typically represent scenes in low dynamic range (LDR), which restricts their use to evenly lit environments and hinders immersive viewing experiences.
Secondly, their reliance on a pinhole camera model, assuming all scene elements are in focus in the input images, presents practical challenges and complicates refocusing during novel-view synthesis.
Addressing these limitations, we present a lightweight method based on 3D Gaussian Splatting that utilizes multi-view LDR images of a scene with varying exposure times, apertures, and focus distances as input to reconstruct a high-dynamic-range (HDR) radiance field.
By incorporating analytical convolutions of Gaussians based on a thin-lens camera model as well as a tonemapping module, our reconstructions enable the rendering of HDR content with flexible refocusing capabilities. 
We demonstrate that our combined treatment of HDR and depth of field facilitates real-time cinematic rendering, outperforming the state of the art.


\end{abstract}  

\section{Introduction}
\label{intro}

Radiance fields~\cite{mildenhall2021nerf} remain the most prevalent representation for novel-view synthesis applications due to their outstanding ability to produce high-quality renderings of scenes from multi-view photos.
3D Gaussian representations~\cite{kerbl20233d} have further advanced the field and now routinely deliver real-time rendering performance.
Most works in this space are concerned with accurate, all-in-focus reconstructions of well-lit scenes.
However, for radiance fields to become versatile tools for storytelling and artistic exploration, \ie to become \textit{cinematic}, they must simultaneously handle high-contrast illumination and depth of field.
To achieve this, we propose a method for reconstructing high-dynamic-range (HDR) radiance fields with real-time re-exposure and re-focusing capabilities based on multi-view low-dynamic-range (LDR) images with varying exposures, apertures, and focus distances.

Considering the significance of the task, naturally, individual aspects of our problem setting have already been investigated in the literature.
On the one hand, the limited dynamic range of typical cameras often leads to overexposed and underexposed regions, causing saturation (clamping to white) and noise, respectively~\cite{reinhard2010high}. 
These limitations directly carry over to radiance fields based on such photos and significantly restrict post-editing capabilities.
Addressing this issue, several works have considered reconstructing HDR radiance fields from multi-view exposure brackets~\cite{huang2022hdr,jun2022hdr,ruckert2022adop} or RAW data~\cite{mildenhall2022nerf}.
On the other hand, defocus blur in input images has been regarded as a degradation to overcome for all-in-focus reconstruction~\cite{wu2022dof,ma2022deblur,lee2023dp,lee2024deblurring}, with synthetic depth of field as an occasional afterthought.
In contrast to all these works, we embrace both HDR and depth of field as expressive tools and set out to develop a radiance field reconstruction approach that enables real-time rendering and high-quality post-editing of exposure, aperture, and focus distance.
 
The input to our method is a set of LDR images that can be captured with a consumer-grade camera.
Different from typical radiance field reconstruction settings, the input images vary not only in extrinsic camera parameters, but also in exposure time, aperture size, and focus distance.
This diverse input sample distribution allows our method to build an all-in-focus HDR 3D Gaussian representation that is optimized for a subsequent application of lens blur.
Depth of field is modeled as a highly efficient convolution of projected scene Gaussians with a Gaussian aperture function~\cite{krivanek2003fast}, based on a thin-lens model.
The convolved Gaussians are then rasterized and projected to LDR using a dedicated tonemapping module.
Once trained, our model enables novel-view synthesis with arbitrary combinations of camera and lens parameters, facilitating flexible post-editing in terms of exposure and depth of field in real time.
We show that our approach can synthesize high-quality and editable imagery, including appealing bokeh effects that require the joint treatment of HDR and depth of field.
We also demonstrate that our system is on par with and often outperforms the state of the art in the individual disciplines of HDR and all-in-focus reconstruction.
This evaluation is based on multiple new datasets we introduce, including both synthetic and real data, all available on our webpage \url{https://cinegs.mpi-inf.mpg.de/}.

Our contributions can be summarized as follows:
\begin{itemize}
\item A unified framework for the reconstruction of HDR radiance fields with depth of field.
\item Powerful post-editing capabilities such as exposure selection and arbitrary re-focusing in real time.
\item State-of-the-art quality in HDR and all-in-focus reconstruction as well as new challenging benchmark datasets.
\end{itemize}

\section{Related Work}
\label{rw}

In this section, we discuss previous work on depth-of-field (DoF) rendering and HDR image reconstruction. We also briefly summarize recent techniques for novel view synthesis, such as Neural Radiance Field (NeRF) and 3D Gaussian Splatting (3DGS), with special emphasis on those that support DoF and HDR effects.

\subsection{Depth-of-Field Rendering}
\label{dof}
Depth-of-field (DoF) is the range within a scene where objects appear in focus, with objects outside this range appearing blurred. This effect plays an important role in photography and cinematography, helping to guide viewer attention and enhance depth perception \cite{demers2004chapter}. 
Computer graphics has long sought to emulate depth-of-field effects of real-world camera systems: Cook et al.\cite{cook1984distributed} employ ray tracing to achieve accurate depth-of-field by casting rays across the lens.
 Haeberli et al.\cite{haeberli1990accumulation} use the accumulation-buffer technique, simulating distributed ray tracing by rendering scenes multiple times from different lens positions and blending them. 
 Levoy et al.\cite{levoy2023light} apply the same concepts in light field rendering. 
 However, the quality obtained with these methods is tied to the number of samples taken, implying a high computational effort to produce artifact-free images. To trade visual accuracy for efficiency, Scofield et al.\cite{scofield1992212} apply a layered technique, segmenting scenes into depth-ordered layers for simplified blurring. This method achieves depth-of-field effects while avoiding blurred edges but struggles with depth-spanning objects.
 Krivanek et al.\cite{krivanek2003fast} propose a surface splatting method suitable for real-time rendering; occasional artifacts are caused by approximate surface reconstructions.
 Potmesil et al.\cite{potmesil1982synthetic} utilize forward mapping to render sprites for depth-of-field effects 
 and blending pixels as circles in the frame buffer, with a final renormalizing pass for producing accurate alpha values. This post-processing method for simulating depth-of-field has influenced many subsequent works \cite{won2022learning,si2023fully,pidhorskyi2022depth,wang2023implicit}, which calculate the circle of confusion (CoC) size based on depth and use spatially varying convolution to perform weighted averaging. 

\subsection{High Dynamic Range Image Reconstruction}
\label{hdr}

Typically, we distinguish three major classes of HDR image construction \cite{reinhard2010high}: specialized HDR cameras, single-shot methods that perform dynamic range expansion for an input low dynamic range (LDR) image, and multi-shot methods that merge differently exposed LDR images into an HDR output.
Specialized HDR cameras have seen limited adoption in practice due to their high costs and sensor technology limitations \cite{eilertsen2018high}.
Single-shot (or \textit{inverse tone mapping}) methods \cite{banterle2017advanced} are an attractive alternative for handling legacy LDR images. 
They easily synergize with machine learning methods \cite{marnerides2018expandnet,endo2017deep,liu2020single,santos2020single} for denoising underexposed regions and inpainting overexposed details \cite{eilertsen2017hdr,wang2023glowgan}.
Nevertheless, the fidelity of these methods falls short of true multi-exposure methods\cite{debevec2023recovering}.
Here, multiple captures are generated at differing exposures. However, merging those exposures for dynamic scenes or handheld cameras is challenging due to ghosting and motion blur\cite{sen2012robust,hu2012exposure}.
Pioneered by Kalantari et al. \cite{kalantari2017deep}, a further boost of reconstructed HDR quality is achieved by learning-based methods. 
Recent research in this area focuses on exploring different network architectures, such as CNNs \cite{kalantari2017deep, yan2019attention, yan2020deep, yan2022lightweight, xiong2021hierarchical} and Transformers \cite{chen2023improving, liu2022ghost, tel2023alignment}.
Further work explores the integration of generative adversarial losses \cite{niu2021hdr}, diffusion priors \cite{yan2023towards}, as well as semi-supervised \cite{yan2023smae, prabhakar2021labeled} and self-supervised methodologies \cite{SelfHDR}.

\subsection{Depth-of-Field and High Dynamic Range Radiance Fields}
\label{rf}
Significant progress has been made in reconstructing radiance fields from 2D images, with NeRF \cite{mildenhall2021nerf} and 3D Gaussian Splatting \cite{kerbl20233d} techniques being particularly noteworthy.
Neural Radiance Fields (NeRF) \cite{mildenhall2021nerf} use a multi-layer perceptron (MLP) with positional encoding to map 3D locations and viewing directions to color and density values.
Parameter optimization is achieved via gradient descent on the photometric error.
Subsequent work on NeRF \cite{hedman2021baking, 
verbin2022ref, barron2021mip, barron2022mip, barron2023zip, 
muller2022instant} focused on further improving its quality and initially slow rendering speed of this implicit representation. 
In contrast, 3D Gaussian Splatting (3DGS) \cite{kerbl20233d} uses explicit 3D Gaussian primitives as a flexible and expressive scene representation.
The optimization starts from point clouds and optimizes per-Gaussian properties. 
While effective, the recency of 3DGS entails minor flaws in its method.
Mip-Splatting \cite{yu2023mip} and multi-scale 3D Gaussian splatting \cite{yan2023multi} both address aliasing artifacts present in the original approach. Mip-Splatting uses 3D size constraints and 2D Mip filters to enhance image quality for zoomed-in or zoomed-out views. Multi-scale 3D Gaussian splatting maintains Gaussians at different scales, using more small Gaussians for high resolution and larger Gaussians for low resolution.
 
In most radiance field research, it is assumed that input images are consistently in focus and well-exposed. 
While many methods can tolerate small deviations from those assumptions, few recent methods more actively model variations in focal length and exposure time.
DoF-NeRF \cite{wu2022dof} extends NeRF to handle shallow depth-of-field inputs and simulate DoF effects through a differentiable circle of confusion representation. Deblur-NeRF \cite{ma2022deblur} improves NeRF by addressing image blurriness from defocus or motion with a Deformable Sparse Kernel (DSK) module, enabling sharp 3D reconstructions. DP-NeRF \cite{lee2023dp} tackles geometric and appearance consistency issues in blurred images using physical priors and adaptive weighting, significantly enhancing 3D reconstruction quality in the presence of motion and defocus blur.
Concurrent work \cite{lee2024deblurring, darmon2024robust} proposes to compensate for defocus blur in captures with Gaussian splats by inferring alternative covariances for out-of-focus Gaussians.
HDR-NeRF \cite{huang2022hdr} recovers HDR radiance fields from LDR views with different exposures, enabling the generation of novel HDR and LDR views by modeling the physical imaging process integrated with NeRF. 
In freshly appearing concurrent work \cite{cai2024hdrgs}, a similar approach has been proposed for 3DGS that greatly reduces training and inference time.
RawNeRF \cite{shade1998layered, mildenhall2022nerf} utilizes a modified NeRF technique that trains directly on linear raw images to preserve the scene’s full dynamic range, enabling high-quality HDR novel view synthesis. It should be noted that RawNeRF uses raw format inputs without defocus blur and requires the multiplane image-based blurring algorithm \cite{kraus2007depth} for refocusing. VR-NeRF~\cite{xu2023vr} trains HDR radiance fields by mapping the full dynamic range to a perceptually linear space, allowing HDR models to be learned with values constrained to the unit range.
In contrast to these works, our approach uses defocused 8-bit LDR inputs, jointly models depth of field and tone mapping processes, and inherently supports all post-editing tasks.
The methods discussed above can handle simple DoF and HDR factors independently. 
However, they cannot account for both simultaneously; this causes issues in casual captures, e.g., scenes that contain details in the near foreground and far background, as well as strong luminance variation between dark shadows and bright lights. 

As shown by TAF \cite{wang2023implicit}, a stack of images with different exposure/aperture/focus can be employed for all-in-focus HDR image reconstruction of particularly demanding scenes for a specific camera pose. 
Our goal is to generalize this approach to multiple camera poses within a novel-view synthesis framework based on 3DGS for efficient rendering.
Capturing a complete image stack for each input camera pose would be impractical; we aim to sample the space of camera parameter setups between neighboring camera poses so that, as in traditional NeRF and 3DGS methods, a single image per pose is effectively captured. 
This way, we enable high-quality reconstruction of scenes with high variation of depth and contrast ranges, and their rendering with arbitrary camera parameters for any view.
Our approach directly benefits from the more realistic depth-of-field simulations when performed using HDR images as the input \cite{zhang2019synthetic}.


\section{Method}
\label{sec:method}

Our approach takes as input a set of LDR multi-view images 
$\{\inputimage_k\}_{k=1}^\numinputimages$ of a scene, captured with randomly selected exposure times \exposuretime, aperture sizes \aperture, and focus distances \focaldistance (Fig.~\ref{fig:teaser}, left).
Since exposure and lens settings are typically stored in EXIF headers, we assume this information is available for each image.
Using this input, our method reconstructs an HDR radiance field that allows for post-editing of exposure and depth of field.

Our approach is based on the 3D Gaussian Splatting representation~\cite{kerbl20233d}, which we briefly review in Sec.~\ref{sec:background}.
The first step in our pipeline is calibrating the input cameras, detailed in Sec.~\ref{sec:cam_calibration}.
We then introduce our model, which features an HDR 3D Gaussian scene representation with an integrated defocus blur simulator (Sec.~\ref{sec:representation}).
A tonemapping module allows us to train the representation on LDR input only (Sec.~\ref{sec:training}).

\begin{figure*}[h!]
    \centering
    \includegraphics[width=\linewidth]{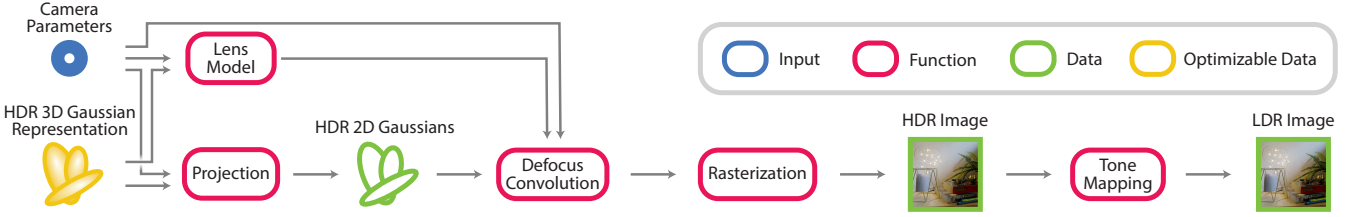}
    \caption{Our pipeline for rendering HDR radiance fields with controllable depth of field.}
    \label{fig:pipeline}
\end{figure*}


\subsection{Background}
\label{sec:background}

Radiance fields based on 3D Gaussians~\cite{kerbl20233d} represent a scene using a Gaussian mixture model.
Each primitive has a mean 
$\mean \in \mathds{R}^3$, 
a symmetric covariance matrix 
$\covariance \in \mathds{R}^{3 \times 3}$, 
an opacity $\opacity \in \mathds{R}$, as well as a set of spherical harmonics (SH) coefficients encoding view-dependent RGB color $\gaussiancolor \in \mathds{R}^3$.
To render an image using this representation~\cite{zwicker2001ewa}, the primitives are projected to 2D screen space using a pinhole camera model to obtain a projected mean 
$\projectedmean \in \mathds{R}^2$ 
and projected covariance
$\projectedcovariance \in \mathds{R}^{2 \times 2}$.
The final pixel color is obtained using front-to-back compositing:
\begin{equation}
\label{eq:rendering}
    \pixelcolor(\pixelposition)
    =
    \sum_{i \in \mathcal{N}(\pixelposition)}
    \gaussiancolor_i
    \gaussianalpha_i(\pixelposition)
    \prod_{j=1}^{i-1}
    \left(
        1 - \gaussianalpha_j(\pixelposition)
    \right),
\end{equation}
where $\mathcal{N}$ is the sequence of depth-ordered primitives overlapping the pixel at location $\pixelposition \in \mathds{R}^2$, and
\begin{equation}
\label{eq:alpha}
    \gaussianalpha_i(\pixelposition)
    =
    \opacity_i
    \cdot
    \exp
    \left(
        - \frac{1}{2}
        (\pixelposition - \projectedmean_i)^T
        \projectedcovariance_i^{-1}
        (\pixelposition - \projectedmean_i)
    \right)
\end{equation}
is the opacity of the primitives.
Since the rendering function in Eq.~\ref{eq:rendering} is differentiable, the model can be trained from posed input images via gradient-based optimization.


\subsection{Camera Calibration}
\label{sec:cam_calibration}

We require posed input images to build our model.
The standard approach for this is Structure from Motion (SfM)~\cite{schonberger2016structure}, which we run on our input data.
Unfortunately, such photometric calibration leaves the scene's global scale ambiguous.
As various lens parameters are provided in absolute millimeters, we need to scale the scene accordingly.
To achieve this, we leverage the defocused images and their respective lens configurations. Specifically, we run an image blur detector~\cite{alireza2017spatially} on all input images to obtain per-pixel sharpness estimates. By thresholding these sharpness maps, we identify the image space locations of in-focus scene elements.
The known focus distance \focaldistance per input image provides ground-truth depth values for these pixels.
Next, we train a vanilla 3D Gaussian Splatting model on the input images after rescaling pixel intensities to a common exposure.
Using this model, we render a depth map for each input view.
As expected, the renderings from this model are of low quality given our inputs, but we find the overall scene configuration precise enough for our calibration.
Using the depth values of the in-focus pixels, we perform a least-squares fit of the scene scale to match the rendered depth values with the ground truth.

After calibration, we have the full set of camera and lens parameters per input view at our disposal. 
Please note that we follow the methodology established in 3DGS ~\cite{kerbl20233d}. The initialization process differs for synthetic and real datasets: for real scenes, we use SfM for camera and scale calibration, and initialize the scene Gaussians using SfM points. For synthetic scenes with predefined camera parameters, SfM is unnecessary, eliminating scale ambiguity and calibration. In this case, scene Gaussians are initialized with random noise (see Sec. ~\ref{sec:imp}).



\subsection{HDR Radiance Field with Depth of Field}
\label{sec:representation}

Here we describe our pipeline, which consists of an HDR 3D Gaussian scene representation, a defocus convolution module, as well as a tonemapper that projects HDR to LDR content.
An overview of our model is shown in Fig.~\ref{fig:pipeline}.

Our HDR Gaussians differ from conventional LDR Gaussians in one crucial property:
the per-primitive SH coefficients encoding view-dependent color \gaussiancolor exhibit a significantly larger dynamic range.
We denote this corresponding HDR color as \gaussiancolorhdr.
Notably, no architectural changes compared to an LDR representation are necessary at this stage; the high dynamic range emerges from our carefully designed pipeline, as detailed in the remainder of this section.

The next stage of our model is concerned with depth of field (DoF) synthesis, which occurs at the level of 2D Gaussians projected into screen space.
Considering a thin-lens model~\cite{potmesil1981lens}, a projected point with depth \depth is spread over a disk -- typically referred to as the circle of confusion -- with radius
\begin{equation}
    \label{eq:coc}
    \coc(\depth) = 
    \frac
        {\left|  \depth - \focaldistance \right| \cdot \focallength^2}
        {2 \cdot \aperture \cdot \depth \cdot \left(\focaldistance - \focallength \right)} \text{,}
\end{equation}
where \focallength is the focal length, \aperture is the F-number, representing the aperture size and \focaldistance is the focus distance.
We seek to enlarge each projected scene Gaussian to reflect the circle of confusion~\cite{krivanek2003fast}, where the mean \mean determines depth \depth.
We approximate the circle of confusion using an aperture Gaussian with covariance
$\aperturegaussian = 
\text{diag}\left(3 \coc(\depth), 3 \coc(\depth)\right) \in \mathds{R}^{2 \times 2}$
to allow for efficient convolutions with the scene Gaussians.
This convolution can be carried out in closed form~\cite{bromiley2003products,celarek2022gaussian}, resulting in 
an update of Eq.~\ref{eq:alpha}:
\begin{equation}
    \gaussianalphadefocus_i(\pixelposition)
    =
    \opacity_i
    \cdot
    \normalizationfactor_i
    \cdot
    \exp
    \left(
        - \frac{1}{2}
        (\pixelposition - \projectedmean_i)^T
        (\projectedcovariance_i + \aperturegaussian_i)^{-1}
        (\pixelposition - \projectedmean_i)
    \right),
\end{equation}
with the energy-preserving normalization factor
\begin{equation}
    \normalizationfactor_i = 
    \sqrt{
        \frac
            {\det(\projectedcovariance_i)}
            {\det(\projectedcovariance_i + \aperturegaussian_i)}
    }.
\end{equation}
Importantly, each projected scene Gaussian is convolved with an aperture Gaussian reflecting the depth of the scene Gaussian and the lens configuration, producing spatially-varying defocus blur.
An HDR image with defocus blur can now be rasterized using the quantities developed above:
\begin{equation}
\label{eq:hdr_rendering}
    \pixelcolordefocushdr(\pixelposition)
    =
    \sum_{i \in \mathcal{N}(\pixelposition)}
    \gaussiancolorhdr_i
    \gaussianalphadefocus_i(\pixelposition)
    \prod_{j=1}^{i-1}
    \left(
        1 - \gaussianalphadefocus_j(\pixelposition)
    \right).
\end{equation}

Naturally, the HDR content rendered in the previous step has an unbounded range of radiance values. 
Both training and final visualization require values in the range $[0, 1]$. Therefore, we project HDR values to LDR using the tone mapping function
\begin{equation}
\label{eq:tm}
    \pixelcolordefocusldr(\pixelposition)
    = 
    \left(
        \min(\frac{\exposuretime}{\aperture^2} \cdot \pixelcolordefocushdr(\pixelposition), 1)
    \right)^{1/\gamma},
\end{equation}
simulating the per-pixel non-linear response of a real-world camera~\cite{allen2011manual}.
Here, the final exposure is determined using exposure time \exposuretime and aperture size \aperture, before the intensities are clipped to the target domain.
Finally, \gammacorrection performs a gamma correction from linear radiance to pixel intensities. We use an explicit tone mapping operator as it minimizes computational overhead and allows convenient metadata incorporation.

Eq.~\ref{eq:tm} produces images that the user can flexibly edit across multiple dimensions:
In addition to free viewpoint selection, the user can adjust the aperture size \aperture, exposure time \exposuretime, and focus distance \focaldistance in real-time, providing a wide range of artistic control.
\subsection{Training}
\label{sec:training}

\begin{figure}
    \centering    \includegraphics[width=\linewidth]{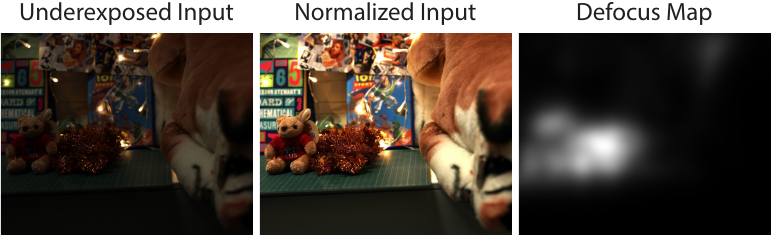}
    \caption{Our training data includes underexposed and overexposed images (\textit{left}). 
    Our loss incorporates both the original data (Eq.~\ref{eq:l_rec}) and a normalized version of it (Eq.~\ref{eq:l_exp}). 
    To manage defocus blur, a third loss component (Eq.~\ref{eq:l_foc}) incorporates a defocus map.
    }
    \label{fig:training}
\end{figure}

We train our model using the loss function
\begin{equation}
    \mathcal{L}
    = 
    \mathcal{L_\text{rec}}
    +
    \lambda_\text{exp}
    \mathcal{L_\text{exp}}
    +
    \lambda_\text{foc}
    \mathcal{L_\text{foc}}.
\end{equation}
All three terms make use of the photometric consistency measure \discrepancy that evaluates a sum of $\ell_1$-distance and DSSIM~\cite{wang2004image} metric.
Yet, each term applies \discrepancy to different versions of the data.
The first term simply measures the consistency between our rendered images and the input data for all pixels \pixelposition and training views $k$:
\begin{equation}
\label{eq:l_rec}
    \mathcal{L_\text{rec}}
    =
    \mathbb E_{\pixelposition, k}
    \left[
    \discrepancy
    \left(
        \pixelcolordefocusldr_k(\pixelposition), \inputimage_k(\pixelposition)
    \right)
    \right].
\end{equation}
To handle the vastly different exposures in our inputs, the second loss term normalizes the images to a medium exposure by scaling pixel intensities before computing discrepancies~\cite{liu2020single, wang2022learning} (Fig.~\ref{fig:training}, left and center):
\begin{equation}
\label{eq:l_exp}
    \mathcal{L_\text{exp}}
    =
    \mathbb E_{\pixelposition, k}
    \left[
    \discrepancy
    \left(
        \frac
            {\pixelcolordefocusldr_k(\pixelposition)}
            {2 \cdot \mathbb E_{\pixelposition}
            \left[
                \pixelcolordefocusldr_k(\pixelposition)
            \right]}
        , 
        \frac
            {\inputimage_k(\pixelposition)}
            {2 \cdot \mathbb E_{\pixelposition}
            \left[
                \inputimage_k(\pixelposition)
            \right]}
    \right)
    \right].
\end{equation}
Finally, due to the prominent occurrence of defocus blur in our inputs, we found it beneficial to include a loss term that explicitly addresses sharpness.
Specifically, we use an off-the-shelf differentiable image blur detector \blurdetector~\cite{alireza2017spatially} that outputs a scalar defocus map (Fig.~\ref{fig:training}, right) and compute
\begin{equation}
\label{eq:l_foc}
    \mathcal{L_\text{foc}}
    =
    \mathbb E_{\pixelposition, k}
    \left[
    \discrepancy
    \left(
        \blurdetector
        \left(
            \pixelcolordefocusldr_k(\pixelposition)
        \right), 
        \blurdetector
        \left(
            \inputimage_k(\pixelposition)
        \right)
    \right)
    \right].
\end{equation}
We set $\lambda_\text{exp} = 0.25$ and $\lambda_\text{foc} = 1$ in all our experiments and train our models using the Adam~\cite{kingma2014adam} optimizer with default parameters, also including all densification strategies from 3DGS~\cite{kerbl20233d}.

\section{Results and Evaluation}
\label{experiments}

In this section, we demonstrate the effectiveness of our method through a comprehensive set of experiments. 
We first describe our experimental settings in Sec.~\ref{sec:exp_setting}, introducing our dataset, which consists of both synthetic and real captured images used for evaluation, followed by the implementation details used for all results.
Then, in Sec.~\ref{sec:es}, we evaluate separately the capability of our model to reconstruct all-in-focus and HDR images, comparing our results to those of specialized state-of-the-art solutions. After this, we present in Sec.~\ref{sec:fullpip} results for our complete pipeline, showcasing HDR radiance field reconstructions as well as post-editing applications.
Lastly, in Sec.~\ref{ab} we provide ablation studies of the components of our approach. Please refer to the \supp{supplementary video} for additional results. 


\subsection{Experimental Settings}
\label{sec:exp_setting}

\subsubsection{Dataset}

Our dataset includes four synthetic rendered scenes and four real captured scenes. The synthetic scenes provide control and ground truth for evaluation, while the real captured scenes demonstrate the practical performance of our approach. A sample of all scenes can be found in the \supp{supplementary material}.

\noindent\textbf{Synthetic dataset} Our synthetic dataset is composed of four diverse indoor scenes rendered with Cycles (path tracer) in Blender 4.1. The scenes encompass challenging scenarios with high dynamic range of illumination (bright light sources and dark regions illuminated by bounced light) and spacious layouts allowing for multiple focal planes (examples shown in Figures \ref{fig:hdr_results} and \ref{fig:results}). All scenes have been rendered with high number of samples (ranging from 4096 to 16384) to prevent undesired distortions in the defocused regions caused by denoising.
For each scene we generated 960 HDR images (8 $\times$ 8 views sampled at a regular grid wrapped to the sphere surface, 5 uniformly sampled focus distances and 3 different apertures). For each view we generated all-in-focus ground truth image. This results in 1024 images per scene. For evaluation purposes we generated another set of 7 $\times$ 7 views located between the training views. All images are accompanied by a file containing parameters used for rendering (camera transformation matrix, focal length, focus distance, aperture, etc.).
In addition to the HDR content we have modified the scenes for the AiF reconstruction comparison, producing pleasant LDR images with reduced dynamic range without clipped regions (examples shown in Figure \ref{fig:defocus_results}). The camera locations, apertures, and focus distances are sampled in a same manner as in the case of the HDR subset.

\noindent\textbf{Captured dataset} Our captured dataset is composed of four real scenes. Some examples are shown in Figs.~\ref{fig:teaser}, \ref{fig:aif-hdr}, and \ref{fig:results}. These scenes were captured using Canon EOS RP and Canon EOS Rebel T6i cameras, outfitted with RF 85mm F2.0, and EF-S 18-55mm F3.5-5.6 lenses. Each scene was captured by moving the camera to follow an approximately 9$\times$9 grid pattern, resulting in 81 images per scene. At each camera position in this grid, we varied the exposure time, aperture, and focus distance to ensure a diverse set of configurations.

\subsubsection{Implementation}
\label{sec:imp}

We use the 3DGS framework~\cite{kerbl20233d} as our baseline, integrating a thin-lens model and a tone mapping module as described in Sec.~\ref{sec:representation}. The implementation is done in PyTorch and CUDA.
All source code is available on our project webpage.

Additionally, we incorporate some practical design choices to tackle our task more effectively.
Given that our synthetic dataset starts from random noise without an SfM point cloud initialization, we employ a coarse-to-fine optimization strategy. During the initial 7,000 iterations, we exclude tone mapping and depth of field simulation, training on an exposure-aligned dataset to generate an initial point cloud. Following this, we refine the optimization using the coarse result as a starting point. Our first stage training adapts well to a broad range of dynamic variations. However, due to sparse viewpoints, we observed instability when the scene's dynamic range exceeds 18 stops.
The iteration process consists of 40,000 steps, taking approximately 21 minutes on an RTX A40 GPU, using a training dataset with a resolution of 1200 × 675.





\subsection{Comparisons}
\label{sec:es}

In this section we evaluate two important features of our method: all-in-focus reconstruction and HDR reconstruction. We compare our method against specialized state-of-the-art approaches designed to address these specific problems. To facilitate this comparison, we leverage our synthetic dataset, which provides access to ground truth images for numerical evaluations.
While these specialized methods excel in their respective domains, it is important to note that they are limited to solving only one of the problems—those optimized for all-in-focus reconstruction cannot handle effectively HDR, and vice versa. Our approach, however, integrates both capabilities and demonstrates comparable, and in some cases superior performance to these specialized solutions. 

\subsubsection{All-in-Focus Reconstruction}

\begin{figure*}
    \centering
    \includegraphics[width=\linewidth]{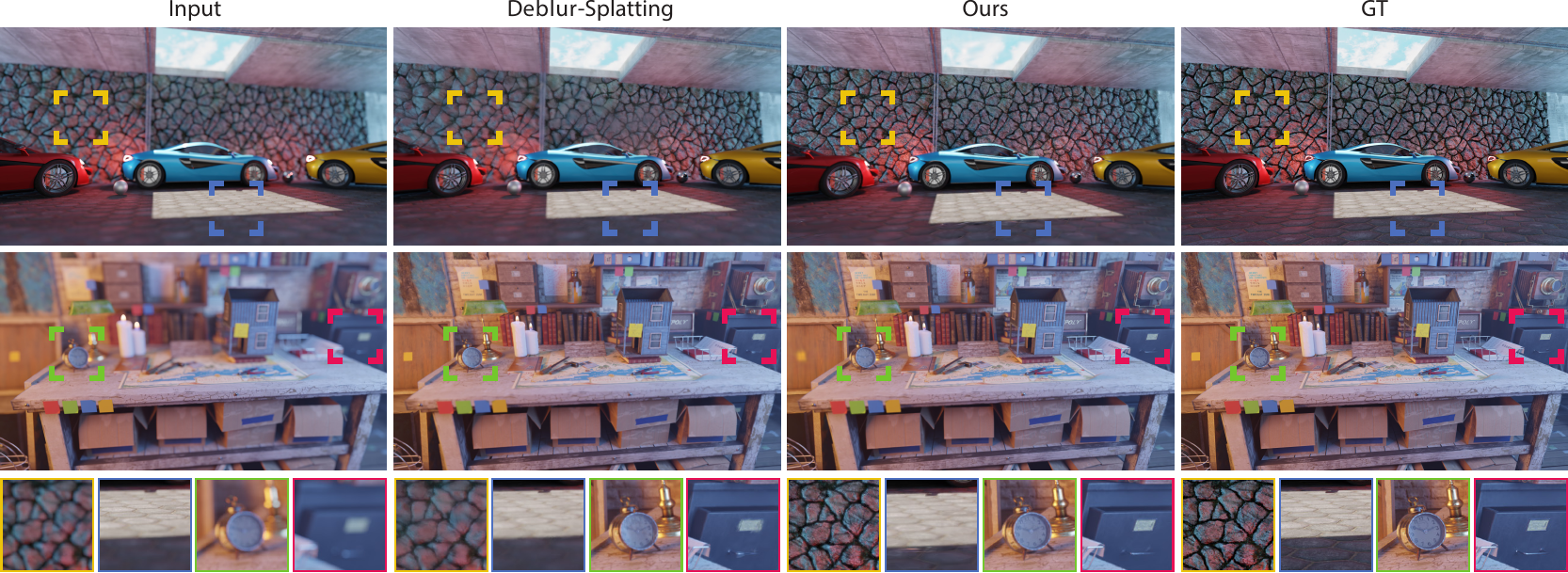}
    \caption{Comparison of all-in-focus reconstruction. Our method provides better or comparable results to Deblur-Splatting method.}
    \label{fig:defocus_results}
\end{figure*}

For the all-in-focus reconstruction task, we compare our results against the concurrent work Deblur-Splatting~\cite{lee2024deblurring}. 
Additionally, we include in the \supp{supplementary material} comparisons for Deblur NeRF~\cite{ma2022deblur}, where we observed noticeable deformations along the boundaries, causing straight lines to appear curved. 
We use the four scenes from our synthetic dataset for this evaluation.
In particular, for each of the scenes, we use 64 (8 $\times$ 8) viewpoints for training, each exhibiting varying degrees of defocus blur, and 49 (7 $\times$ 7) novel viewpoints not seen during training for the evaluation. 
Table~\ref{tab:aif_reconstruction} presents the numerical results for PSNR, SSIM~\cite{wang2004image}, and LPIPS~\cite{zhang2018perceptual}, commonly used for evaluating image quality. 
Additionally, Fig.~\ref{fig:defocus_results} shows illustrative comparisons to the ground truth. 
Although our method is not specifically designed for deblurring, it achieves results comparable to state-of-the-art method. Regarding rendering time, both methods are similarly efficient, achieving approximately 110 frames per second (FPS) on an RTX A40 GPU. We primarily evaluate the synthetic dataset due to the challenge of obtaining ground truth for real datasets. To validate our method, we run Deblur-Splatting on our real dataset using non-reference metrics. Since Deblur-Splatting cannot generate HDR, we tone-map our HDR images and evaluate in the LDR domain using NIQE \cite{mittal2012making} and BRISQUE \cite{mittal2012no}. The results in Table \ref{tab:aif_real} demonstrate our method's superior performance.

\begin{table}[h]
\label{tb:2}
\small
\centering
\caption{Comparison of all-in-focus reconstruction performance metrics for our method and Deblur-Splatting using our synthetic dataset.}
\renewcommand{\tabcolsep}{0.2cm}
\label{tab:aif_reconstruction}

\begin{tabular}{llrrr} 
Scene & Method & PSNR $\uparrow$ & SSIM $\uparrow$ & LPIPS $\downarrow$ \\
\midrule

\multirow{2}{*}{Car} & Deblur-Splatting & 27.23 $\pm$ 3.21 & 0.86 $\pm$ 0.07 & 0.18 $\pm$ 0.04 \\
& \textbf{Ours} & \textbf{28.05 $\pm$ 0.26} & \textbf{0.89 $\pm$ 0.01} & \textbf{0.11 $\pm$ 0.01} \\
\midrule

\multirow{2}{*}{Attic} & Deblur-Splatting & 31.86 $\pm$ 0.23 & 0.91 $\pm$ 0.01 & 0.13 $\pm$ 0.01 \\
& \textbf{Ours} & \textbf{32.90 $\pm$ 0.18} & \textbf{0.93 $\pm$ 0.00} & \textbf{0.09 $\pm$ 0.00} \\
\midrule

\multirow{2}{*}{Café} & Deblur-Splatting & 30.24
 $\pm$ 0.42 & \textbf{0.87 $\pm$ 0.01} & \textbf{0.12 $\pm$ 0.01} \\
& \textbf{Ours} & \textbf{30.50 $\pm$ 0.36} & 0.85 $\pm$ 0.01 & 0.14 $\pm$ 0.00 \\
\midrule

\multirow{2}{*}{Office} & Deblur-Splatting & \textbf{34.59 $\pm$ 0.75} & \textbf{0.95 $\pm$ 0.01} & 0.08 $\pm$ 0.01 \\
& \textbf{Ours} & 31.70 $\pm$ 0.80 & 0.92 $\pm$ 0.01 & \textbf{0.05 $\pm$ 0.01} \\

        \bottomrule
    \end{tabular}
\end{table}

\subsubsection{HDR Reconstruction}

\begin{figure*}
    \centering
    \includegraphics[width=\linewidth]{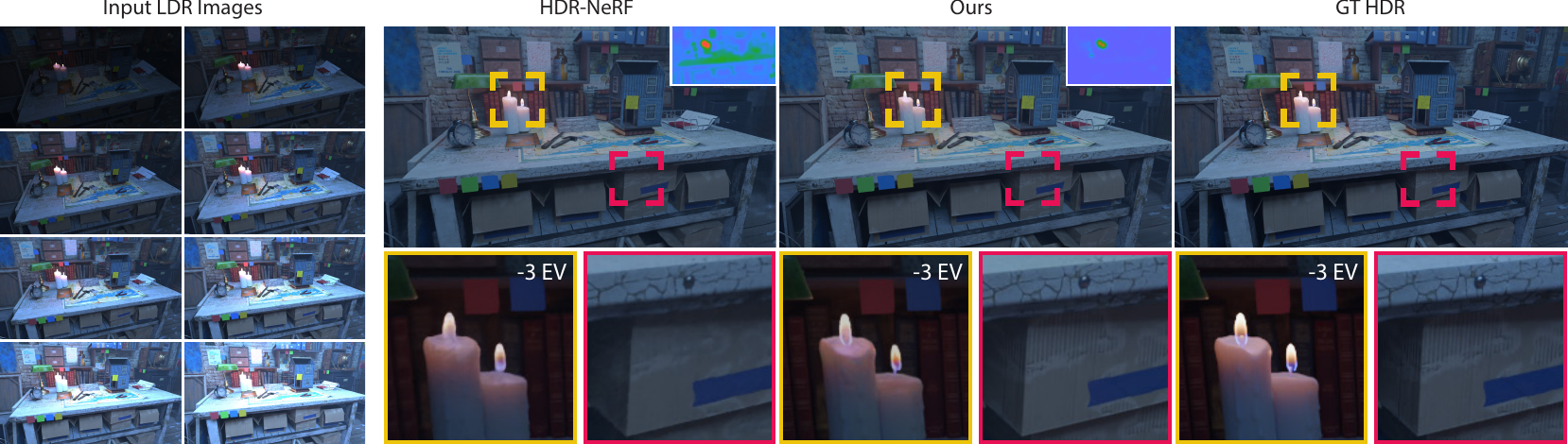}
    \caption{Comparison of HDR reconstruction. Left-most column presents input LDR images for a given view. The following columns depict results acquired by HDR-NeRF, Ground Truth, and results of our method, respectively. Please note, that our method properly reconstructs the shape of the flame and reproduces successfully high-frequency content, while these details are missing in the results generated by HDR-NeRF. To depict the accuracy of both methods we provide the HDR-VDP-3 error map in the top right corner of the images. To better show the differences in the highlights, we render these regions at -3 stops, as shown in the yellow box.}
    \label{fig:hdr_results}
\end{figure*}

For HDR reconstruction, we compare our results against HDR-NeRF~\cite{huang2022hdr} using various metrics specifically targeted towards evaluating HDR quality, including HDR-VDP-3~\cite{mantiuk2023hdr}, PU-PSNR, and PU-SSIM~\cite{azimi2021pu21}, as summarized in Table~\ref{tab:quant_reconstruction}. Visual comparisons with the ground truth are depicted in Fig.~\ref{fig:hdr_results}.
Again, we use for evaluation the four scenes of our synthetic dataset, with 64 (8 $\times$ 8) views for training and 49 (7 $\times$ 7) novel ones for testing. To simulate LDR training views with varying exposures, starting from the HDR renders from our synthetic dataset, we sample a single exposure for each view. For this, we first determine the start and end boundaries that emcompass the entire dynamic range of the scene~\cite{andersson2021visualizing}. Then, we divide this range into eight intervals, within which we sample exposures as described in previous work~\cite{huang2022hdr}.
Our method consistently outperforms HDR-NeRF in almost all metrics, demonstrating superior reconstruction capabilities. Notably, we observed that HDR-NeRF tends to overfit to the LDR images, resulting in lower quality HDR image generation. Moreover, our approach achieves real-time rendering capabilities, significantly outperforming HDR-NeRF. We deliver approximately 110 FPS at a resolution of 1200 x 675, in contrast to HDR-NeRF's rate of approximately 0.021 FPS for rendering a single image, both using a GPU RTX A40. We also run HDR-NeRF on our real dataset using non-reference metrics PU-PIQE \cite{hanji2022comparison} to assess results from novel views (Table \ref{tab:hdr_real}). These results highlight that our method outperforms task-specific approaches.

\begin{table}[h]
\centering
\caption{Comparison of all-in-focus reconstruction performance
metrics for our method and Deblur-
Splatting using our real dataset.}
\label{tab:aif_real}
\begin{tabular}{lcc}
Method & NIQE $\downarrow$ &BRISQUE $\downarrow$ \\
\midrule
Deblur-Splatting & 6.65 $\pm$ 1.87 & 45.77 $\pm$ 8.50 \\
\textbf{Ours} & \textbf{2.66 $\pm$ 0.22} & \textbf{32.17 $\pm$ 4.75} \\
\bottomrule
\end{tabular}
\end{table}


\begin{table}[h]
\label{tb:1}
\small
    \centering
    \caption{Comparison of HDR reconstruction performance metrics for our method and HDR-NeRF using our synthetic dataset.}
    \renewcommand{\tabcolsep}{0.2cm}
    \label{tab:quant_reconstruction}
    
    \begin{tabular}{llrrr} 
    Scene & Method & HDR-VDP-3 $\uparrow$ & PU-PSNR $\uparrow$ & PU-SSIM $\uparrow$ \\
    \midrule

\multirow{2}{*}{Car} & HDR-NeRF & 9.70 $\pm$ 0.15 & \textbf{29.78 $\pm$ 1.72} & 0.91 $\pm$ 0.02 \\
& \textbf{Ours} & \textbf{9.75 $\pm$ 0.09} & 29.67 $\pm$ 4.63 & \textbf{0.95 $\pm$ 0.01} \\
\midrule

\multirow{2}{*}{Attic} & HDR-NeRF & 9.92 $\pm$ 0.03 & \textbf{39.04 $\pm$ 0.71} & 0.96 $\pm$ 0.04 \\
& \textbf{Ours} & \textbf{9.98 $\pm$ 0.01} & 38.97 $\pm$ 1.01 & \textbf{0.98 $\pm$ 0.00} \\
\midrule

\multirow{2}{*}{Café} & HDR-NeRF & 9.16 $\pm$ 0.22 & 26.97 $\pm$ 2.56 & 0.76 $\pm$ 0.11 \\
& \textbf{Ours} & \textbf{9.93 $\pm$ 0.01} & \textbf{30.02 $\pm$ 0.51} & \textbf{0.91 $\pm$ 0.00} \\
\midrule

\multirow{2}{*}{Office} & HDR-NeRF & 9.36 $\pm$ 0.18 & 25.13 $\pm$ 1.97 & \textbf{0.90 $\pm$ 0.01} \\
& \textbf{Ours} & \textbf{9.50 $\pm$ 0.09} & \textbf{30.11 $\pm$ 0.22} & 0.71 $\pm$ 0.05 \\

        \bottomrule
    \end{tabular}
\end{table}

\begin{table}[h]
\centering
\caption{Comparison of HDR reconstruction performance
metric for our method and HDR-NeRF using our real dataset.}
\label{tab:hdr_real}
\begin{tabular}{lc}
Method & PU-PIQE $\downarrow$  \\
\midrule
HDR-NeRF & 55.47 $\pm$ 2.22  \\
\textbf{Ours} &\textbf{42.22 $\pm$ 1.22}  \\
\bottomrule
\end{tabular}
\end{table}

\subsection{HDR Radiance Field Reconstruction and Post-Editing}
\label{sec:fullpip}
\begin{figure*}
    \centering
    \includegraphics[width=\linewidth]{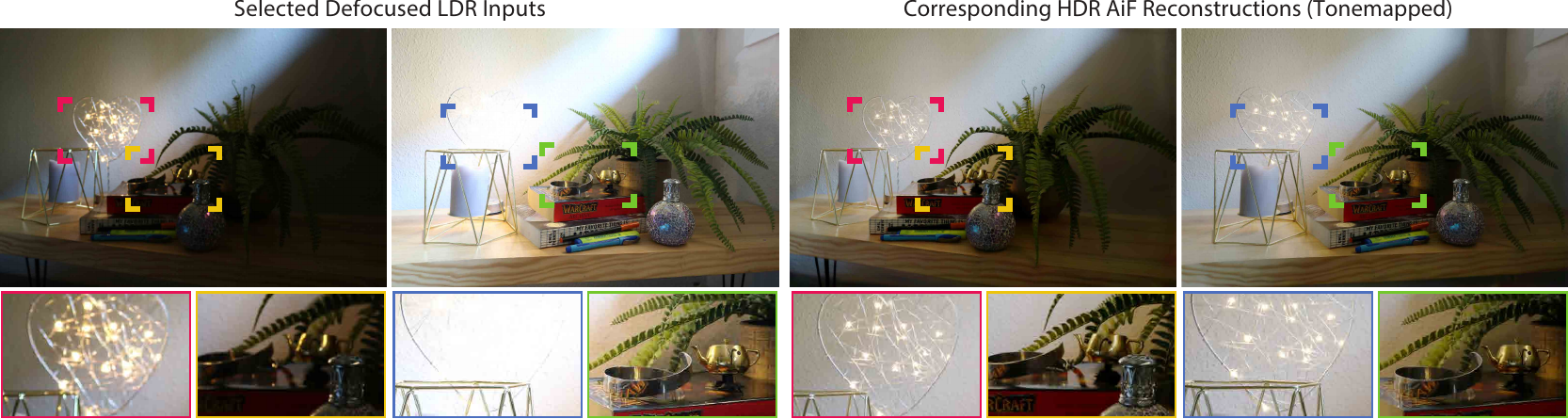}
    \caption{All-in-focus HDR reconstruction. Our method can successfully reconstruct sharp HDR images from defocused images captured with different exposure times. }
    \label{fig:aif-hdr}
\end{figure*}

\begin{figure*}
    \centering    \includegraphics[width=1\linewidth]{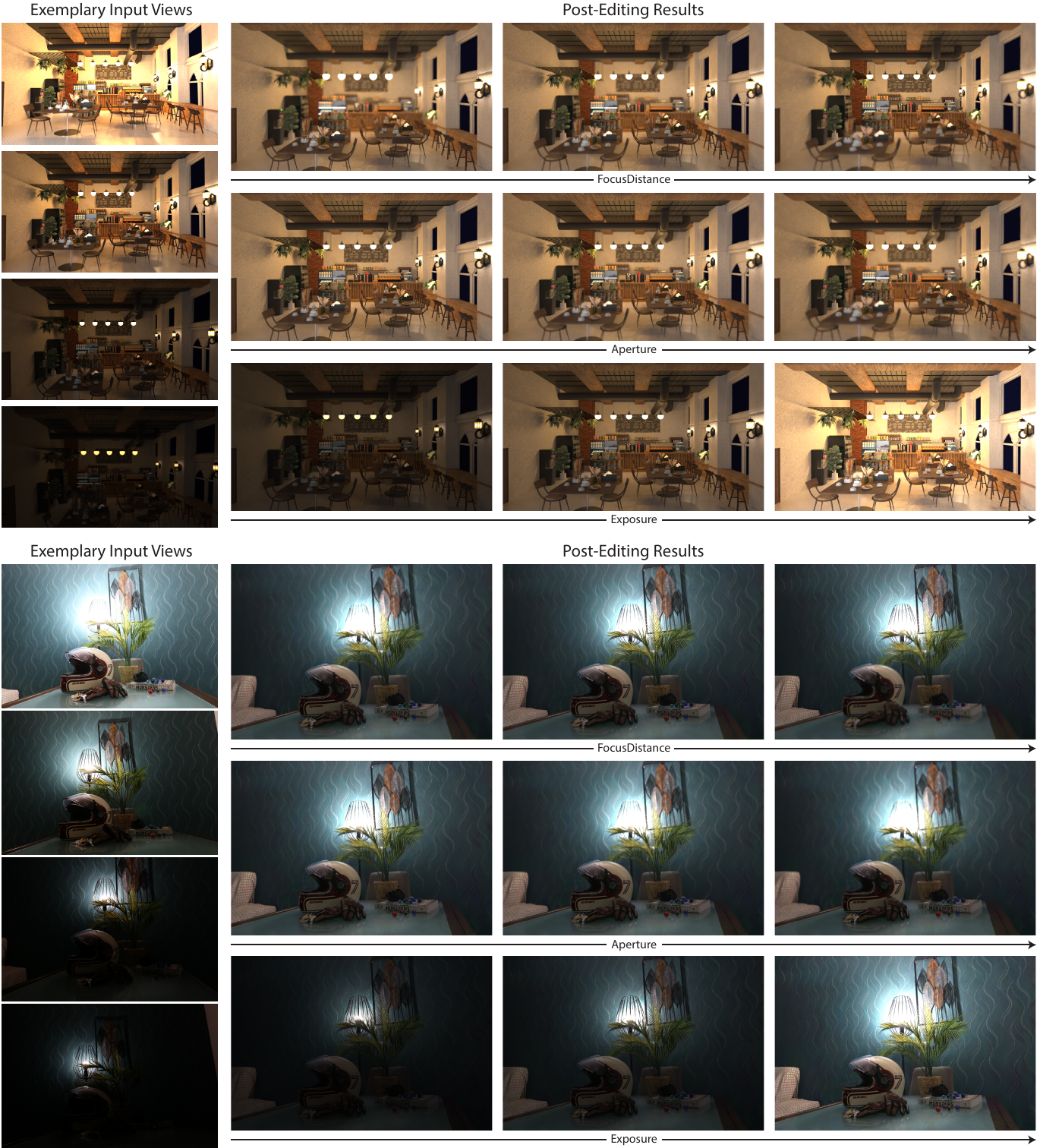}
    \caption{Post-editing results. The leftmost column shows the selected input views. The 3 $\times$ 3 grid to the right displays novel views, with each row illustrating the manipulation capabilities of one parameter: focus distance, aperture size, and exposure.}
    \label{fig:results}
\end{figure*}

In this section, we showcase the full capabilities of our model, specifically its ability to reconstruct HDR radiance fields, which enable the generation of all-in-focus HDR images. 
Fig.~\ref{fig:aif-hdr} presents results from various viewpoints, illustrating how our approach effectively reconstructs detailed scenes with consistent sharpness and dynamic range from multi-view LDR defocused inputs. 
Furthermore, we demonstrate the post-editing capabilities enabled by our method. Our proposed approach reconstructs HDR radiance fields with real-time re-exposure and re-focusing capabilities. Users have the flexibility to adjust aperture size, exposure time, and focus distance in real-time, as demonstrated in Fig.~\ref{fig:results}.


\subsection{Ablation Study}
\label{ab}

We evaluate the impact of the defocus and exposure loss terms by conducting ablation studies on both our real and synthetic datasets.
Additionally, we include in the \supp{supplementary material} an evaluation of our coarse-to-fine training strategy for the synthetic dataset.

\begin{figure}
    \centering
    \includegraphics[width=\linewidth]{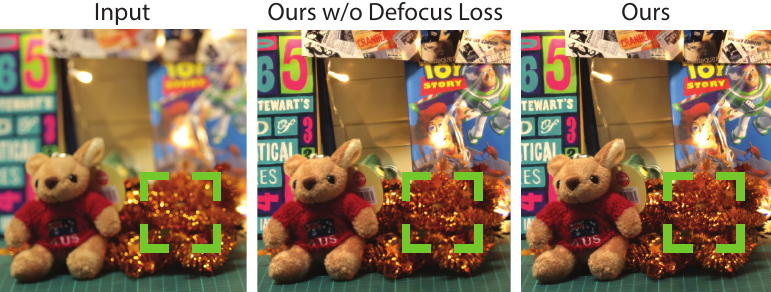}
    \caption{The defocus loss provides higher quality reconstruction for high-frequency patterns.}
    \label{fig:ab_defocus}
\end{figure}

\begin{figure}
    \centering
    \includegraphics[width=\linewidth]{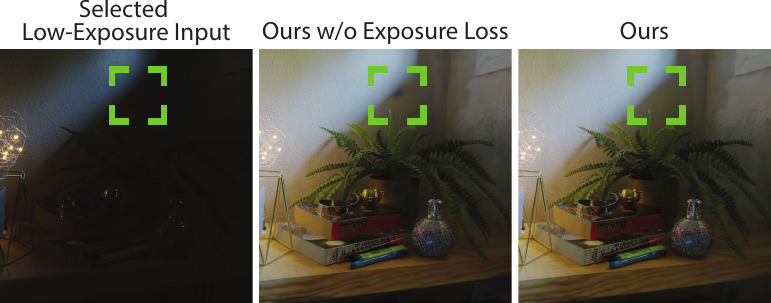}
    \caption{The exposure loss helps to resolve artifacts in the underexposed regions, as the one depicted in the green frame.}
    \label{fig:ab_exp}
\end{figure}

We conduct this evaluation differently for our synthetic and real datasets.
For the synthetic dataset, where ground-truth images are available, we evaluate our all-in-focus HDR reconstructions against ground-truth views not included in the training set. Specifically, we utilize 64 (8 $\times$ 8) views for training and 49 (7 $\times$ 7) novel views for testing. This comparison is performed in HDR space, therefore we use PU-PSNR, PU-SSIM and HDR-VDP-3 as metrics. We show these results in Table~\ref{tab:ab_rendering}.
On the other hand, for the captured dataset, where ground truth is not available, we adopt a different approach. We use 74 captured views for training and leave out 7 for testing. 
Then, we synthesize the testing views with our model using the original aperture, focus distance, and exposure settings corresponding to each captured view, and compare these synthesized views against the actual captures.
This comparison is conducted in LDR space, thus we employ traditional image quality metrics, in particular, PSNR, SSIM, and LPIPS. Quantitative results are presented in Table~\ref{tab:ab_real}, while qualitative results are illustrated in Fig.~\ref{fig:ab_defocus} for the defocus loss term and in Fig.~\ref{fig:ab_exp} for the exposure loss term.


\begin{table}[h]
\centering
\caption{Evaluation of the impact of the defocus and exposure loss terms in the synthetic dataset.}
\label{tab:ab_rendering}
\begin{tabular}{lccc}
Method & PU-PSNR $\uparrow$ & PU-SSIM $\uparrow$ &  HDR-VDP $\uparrow$\\
\midrule
w/o $\mathcal{L}_\text{exp}$ &33.95 $\pm$  1.59 & 0.94 $\pm$  0.00 & 9.80 $\pm$  0.06 \\
w/o $\mathcal{L}_\text{foc}$ &33.63 $\pm$ 1.54 & 0.94 $\pm$  0.00 & 9.78 $\pm$  0.05 \\
Ours &\textbf{34.04 $\pm$ 1.67} & \textbf{0.95 $\pm$ 0.00} & \textbf{9.83 $\pm$  0.05} \\
\bottomrule
\end{tabular}
\end{table}

\begin{table}[h]
\centering
\caption{Evaluation of the impact of the defocus and exposure loss terms in the real dataset.}
\label{tab:ab_real}
\begin{tabular}{lccc}
Methods & PSNR $\uparrow$ & SSIM $\uparrow$ &  LPIPS $\downarrow$\\
\midrule
w/o $\mathcal{L}_\text{exp}$ &23.95 $\pm$ 3.37  & 0.82 $\pm$  0.10 & 0.11 $\pm$  0.04 \\
w/o $\mathcal{L}_\text{foc}$ &24.35 $\pm$ 1.06 & 0.85 $\pm$  0.06 & 0.11 $\pm$  0.03 \\
Ours &\textbf{25.69 $\pm$ 2.71} & \textbf{0.86 $\pm$ 0.07} & \textbf{0.10 $\pm$  0.02} \\
\bottomrule
\end{tabular}
\end{table}

\subsection{Discussion of Depth-of-Field Generation}
\label{subsec:dof_com}
Our depth-of-field module can achieve real-time performance at high quality.
Alternatives include accumulation buffering \cite{haeberli1990accumulation}, which delivers high-quality results but is slow due to the need for multiple samples. Another method, used in RawNeRF \cite{mildenhall2022nerf}, decomposes the image into multiplane images at different depths \cite{shade1998layered, kraus2007depth}, applies image space blurring, and then composites them. While this improves speed, it struggles with large depth complexity. In contrast, our approach does not have this limitation.

\begin{figure}
    \centering
    \includegraphics[width=\linewidth]{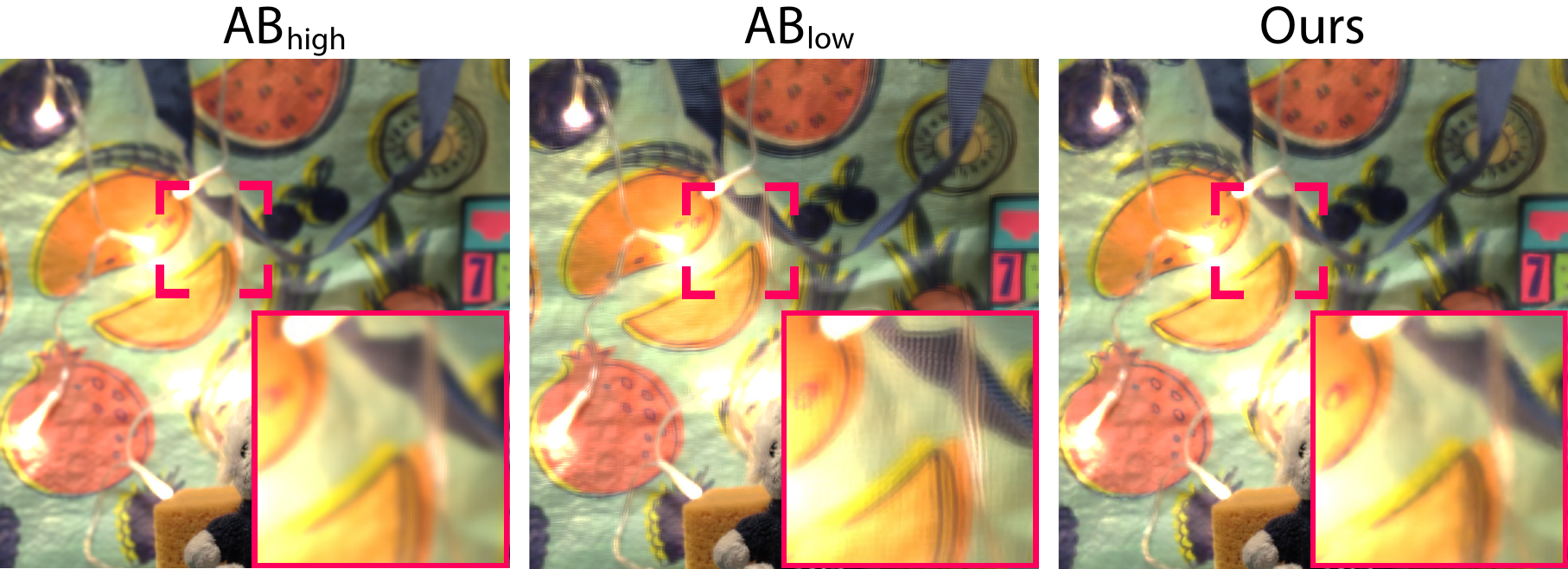}
    \caption{Depth-of-field rendering comparison. Our real-time method achieves high-quality results comparable to those using an accumulation buffer with many samples.}
    \label{fig:dof_com}
\end{figure}

To shed some light on the respective trade-offs, we compare our method against accumulation buffering for depth of field simulation. 
Specifically, we replace our DoF module with accumulation buffering, testing with 20 ($\text{AB}_\text{low}$) and 120 samples ($\text{AB}_\text{high}$). Unlike accumulation buffering, which depends on Monte Carlo sampling, our method requires only a single rendering. As shown in Table ~\ref{tab:1}, our method delivers depth-of-field quality comparable to accumulation buffering with a high sample count while significantly improving speed, enabling real-time rendering. Visual comparisons in Fig. \ref{fig:dof_com} reveal that lower sampling in accumulation buffering results in noticeable artifacts.

\begin{table}[h]
\centering
\caption{Comparisons of our depth-of-field rendering module against solutions involving an accumulation buffer.}
\label{tab:1}
\begin{tabular}{lccc}
Method &PSNR $\uparrow$ & SSIM $\uparrow$ &  FPS $\uparrow$\\
\midrule
$\text{AB}_\text{high}$ & 27.59 $\pm$ 3.39 & 0.94 $\pm$ 0.02 & 1 \\
$\text{AB}_\text{low}$ &27.29 $\pm$ 3.34  & 0.93 $\pm$  0.02 & 6 \\
Ours &27.41 $\pm$ 3.30 & 0.94 $\pm$  0.02 & 90 \\
\bottomrule
\end{tabular}
\end{table}

\section{Limitations and Future Work}
\label{limitations_future_work}
Our approach does not come without limitations, opening up fruitful avenues for future work.
Our defocus model is a two-fold approximation of the actual physical processes in a real camera lens.
First, we use a thin-lens model~\cite{potmesil1981lens}, which, despite being commonly used, is a simplified representation of defocus. 
More sophisticated models have been developed in the forward-rendering literature~\cite{kolb1995realistic} and related work on defocus modeling in the context of reconstruction from image stacks~\cite{wang2023implicit} has shown that advanced effects, such as lens breathing, can significantly improve accuracy.
Second, we approximate the thin-lens circle of confusion with a Gaussian function to achieve real-time rendering performance. While this is a commonly used design decision~\cite{favaro2010recovering,si2023fully}, it results in a slight softening of bokeh due to the deviation from a disk kernel.
We believe that moving towards more realistic lens models has the potential to further increase result quality and cinematic appeal.

\begin{figure}
    \centering    
    \includegraphics[width=\linewidth]{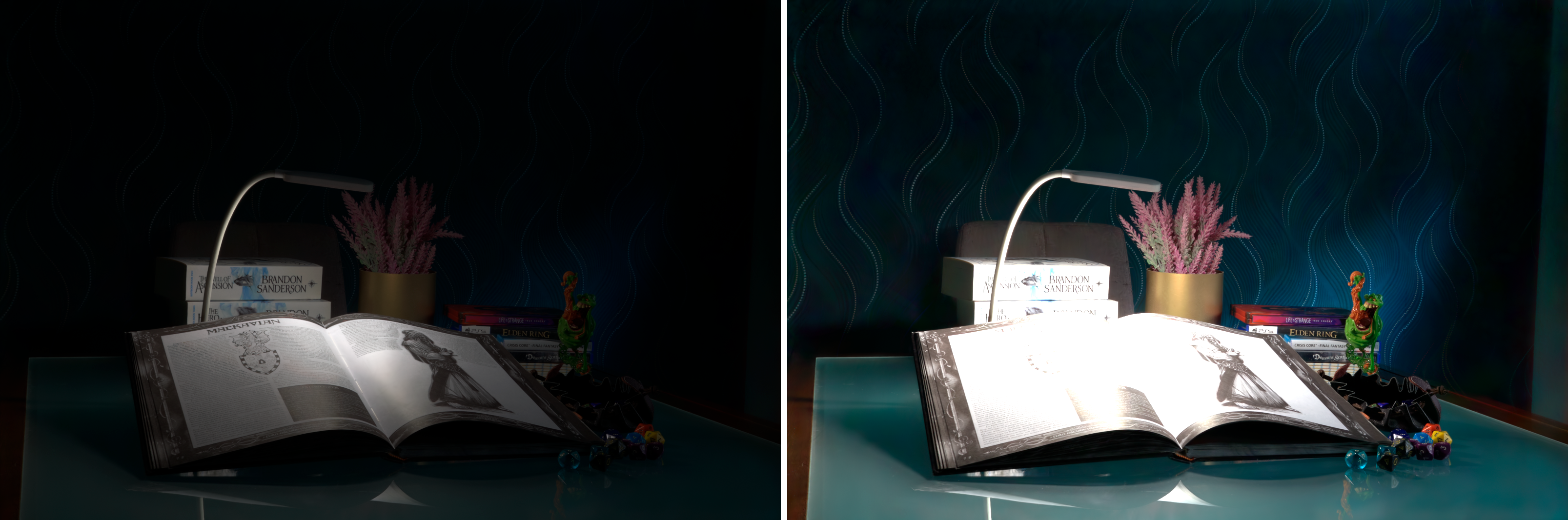}
    \caption{
    Limitation. All-in-focus HDR reconstruction of a scene view with extremely high dynamic range, featuring tone-mapped low (\textit{left}) and high (\textit{right}) exposures. In regions with pronounced darkness and notable noise in the input images, such as the rear wall of the room, artifacts may arise due to information loss. 
    }
    \label{fig:limitation}
\end{figure}

In scenes characterized by extremely high dynamic range, where very dark regions may contain significant noise across most input images, artifacts may occur in the reconstruction of these dark areas.
These artifacts are pronounced when selecting very high exposures (Fig.~\ref{fig:limitation}).
This is expected, since there is insufficient useful information, with noise dominating instead in these areas. 
Nevertheless, these artifacts are noticeable only under extreme exposure settings in the reconstruction, while our method remains capable of delivering pleasing results for the rest of the scene. 
As discussed above, reduced exposure, whether through shorter exposure times or a smaller aperture, results in increased noise in real camera images. However, this paper primarily focuses on HDR and depth of field characteristics. Addressing noise in dark regions is identified as a potential area for future research. Therefore, for our synthetic dataset, we adhere to the HDR-Nerf \cite{huang2022hdr} procedure to render noise-free images.

\section{Conclusion}
\label{conclusion}

We have presented an approach for reconstructing HDR radiance fields with depth of field.
Unlike previous works that treat only exposure variations or defocus blur as a degradation to overcome, we recognize the significance of exposure selection and depth of field for crafting cinematic imagery.
Our pipeline mirrors the physical processes of a camera, enabling us to train a radiance field model using LDR images with various exposure and lens configurations.
We found that our method not only allows powerful and appealing post-editing in real time, but also achieves state-of-the-art quality for the isolated tasks of HDR and all-in-focus reconstruction.
We hope that our approach moves radiance fields a step closer to becoming expressive tools for artists and visual media producers.

\section*{Acknowledgements}

This work has been partially supported by grant PID2022-141539NB-I00, funded by MICIU/AEI/10.13039/501100011033 and by ERDF, EU.

\bibliographystyle{eg-alpha-doi} 
\bibliography{pgbib}

\end{document}